\documentclass[runningheads]{llncs}
\usepackage{graphicx}
\usepackage{comment}
\usepackage{amsmath,amssymb} 
\usepackage{color}

\usepackage{cleveref}
\usepackage{floatrow}

\newcommand{\rf}[1]{\textcolor{blue}{#1}}
\newcommand{\rff}[1]{\textcolor{green}{#1}}

\newcommand{\diego}[2]{\textcolor{black}{#2}}

\begin{document}
\pagestyle{headings}
\mainmatter
\def\ACCV20SubNumber{959}  

\title{Contextual Semantic Interpretability} 

\titlerunning{Contextual Semantic Interpretability}
%
\author{Diego Marcos\inst{1}
\and
Ruth Fong\inst{2}
\and
Sylvain Lobry\inst{1}
\and
R\'emi Flamary\inst{3}
\thanks{Partially funded through the project OATMIL ANR-17-CE23-0012 and 3IA Cote d’Azur Investments ANR-19-P3IA-0002 of the French National Research Agency.}
\and
\newline
Nicolas Courty\inst{4}
\and
Devis Tuia\inst{1,5}}
\authorrunning{D. Marcos et al.}
%

\institute{Wageningen University, The Netherlands \and
Oxford University, UK \and
CMAP, École Polytechnique, Palaiseau, France \and
IRISA, University Bretagne Sud, CNRS, France \and EPFL, Switzerland
}


\maketitle

\begin{abstract}
Convolutional neural networks (CNN) are known to learn an image representation that captures concepts relevant to the task, but do so in an implicit way that hampers model interpretability.
However, one could argue that such a representation is hidden in the neurons and can be made explicit by teaching the model to recognize semantically interpretable attributes that are present in the scene. We call such an intermediate layer a \emph{semantic bottleneck}.
Once the attributes are learned, they can be re-combined to reach the final decision and provide both an accurate prediction and an explicit reasoning behind the CNN decision.
In this paper, we look into semantic bottlenecks that capture \emph{context}: we want attributes to be in groups of a few meaningful elements and participate jointly to the final decision.
We use a two-layer semantic bottleneck that gathers attributes into interpretable, sparse groups, allowing them contribute differently to the final output depending on the context.
We test our contextual semantic interpretable bottleneck (CSIB) on the task of landscape scenicness estimation and train the semantic interpretable bottleneck using an auxiliary database (SUN Attributes).
Our model yields in predictions as accurate as a non-interpretable baseline when applied to a real-world test set of Flickr images, all while providing clear and interpretable explanations for each prediction.
\end{abstract}

\section{Introduction}
Deep learning, in particular convolutional neural networks (CNNs), is increasingly being applied to important yet sensitive domains, such as autonomous driving, facial recognition, and medical applications.
One significant driver behind the success of CNNs is their capacity to learn to approximate complex functions from large amount of data by automatically tuning millions of parameters.
However, this power comes at the expense of interpretability: because of the complexity of CNNs, their internal reasoning can not be easily assessed by humans.
This has implications on scientific and societal levels.

\begin{figure}[!t]
\centering
\floatbox[{\capbeside\thisfloatsetup{capbesideposition={right,top},capbesidewidth=0.4\textwidth}}]{figure}[\FBwidth]
{
\caption{We learn contextual groupings (colored coded) of semantic attributes (middle icons) in order to make predictions (right) (e.g., the meaning of ``road'' depends on the presence of other attributes).}
\label{fig:splash}}
{\includegraphics[width=0.55\textwidth]{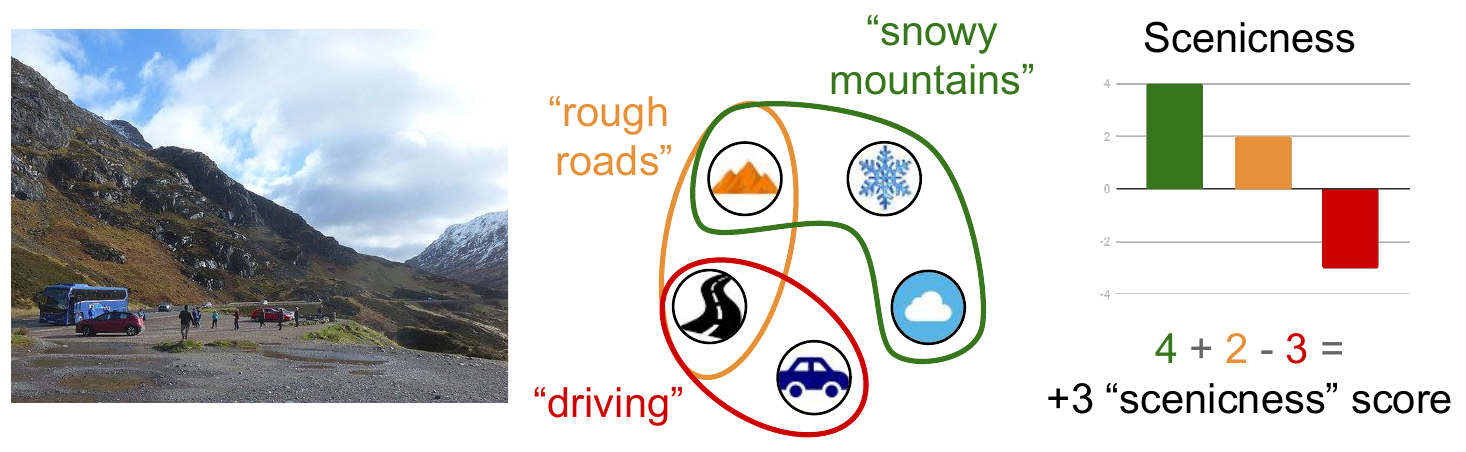}}
\end{figure}

The highly parameterized nature of CNNs enables them to solve a given task in a variety of ways.
Some of these solutions might rely on spurious cues that would harm generalization~\cite{lapuschkin2019unmasking}.
This is well illustrated by numerous works on adversarial examples~\cite{szegedy2013intriguing,kurakin2016adversarial}, in which small perturbations, imperceptible to the human eye, are added to an image and subsequently cause a model to fail.
Furthermore, one can easily find thousands of natural images on which CNNs fail (\emph{e.g.}, real-world adversarial examples)~\cite{hendrycks2019natural}.
Together, these findings cast doubt on the decision functions learned by CNNs and motivate the need for models that are more transparent in their decision-making process.

As deep learning (and its promise of efficient automation) increasingly affects various aspects of human life, the impenetrable complexity of this class of models also becomes a pressing societal issue.
For instance, some governmental entities introduced bills for the regulation of decisions based on algorithms (\emph{e.g.} Equal Credit Opportunity Act in the USA, General Data Protection Regulation in the EU).
While research focused on understanding deep learning predictions is on the rise (see~\cref{sec:SoA}),~\cite{edwards2017slave} highlights that there still is a gap between the understanding of explanability of the machine learning community and that of lawmakers.
Explanations are one way to achieve a degree of interpretability, which can generally be defined as follows: ``systems are interpretable if their operations can be understood by a human''~\cite{biran2017explanation}.
Moreover, the majority of methods that aim to elucidate CNN decisions generate an explanation \emph{a posteriori}; this might induce the risk of a false sense of transparency and trustworthiness~\cite{rudin2019stop}.

In this paper, we make three main contributions.
First, we introduce a novel, explicitly interpretable architecture and training paradigm.
Our model first learns to predict an intermediate, semantically explicit task (e.g., predicting attributes).
These intermediate attributes are then used to make the final prediction on a downstream task.
In particular, we do this by learning a new, sparse, and easily interpretable grouping layer that allows attributes to interact with each other.
We call our proposed layer a Contextual Semantic Interpretable Bottleneck (CSIB).
Second, we demonstrate the interpretability of our model via a novel combination of visualizations.
Using Sankey plots, we visualize both task-specific groups of attributes learned by our model as well as instance-specific explanations that quantify how each attribute (and group) contributed to a model's final prediction.
We also highlight the image regions each group captures.
Together with details on how much each group contributes to the final score, we are able to visualize with clarity, fidelity, and depth the model's decision-making process.
Third, we perform a thorough, empirical analysis of our method applied to the task of scenicness estimation.
Here, we demonstrate that our model performs comparably to a baseline CNN when evaluating a real-world set of Flickr images (there is a performance gap when evaluating on a held-out set from the training distribution).
Lastly, we show our paradigm uniquely allows us to identify and explain systematic errors our model makes.

\section{Related Work}
\label{sec:SoA}

\paragraph{Post-hoc interpretability.}
Most interpretability research introduces post-hoc methods that aim to explain any black-box model (see~\cite{gilpin2018explaining} for an interpretability survey).
Much attention has been focused on the problem of attribution, \emph{i.e.} the identification of image regions (via heatmaps) that are responsible for a model's output~\cite{simonyan14deep,springenberg2014striving,zhou2016learning,selvaraju2017grad,zhang2016excitation,bach2015pixel,zeiler2014visualizing,ribeiro2016should,fong17interpretable,Petsiuk2018rise,fong19understanding}.
Although attribution methods can be applied to any model, the produced heatmaps lack richness, as they only highlight which image regions are decision relevant, but are unable to characterize a specific semantic reasoning or how those image regions interact.
Depending on their formulation, they can also be misleading, as~\cite{adebayo2018sanity} highlights.
Another line of research focuses on understanding the global properties of CNNs.
One approach to this problem is to study CNNs in a scientific manner, \emph{i.e.} by generating and testing hypotheses about CNN properties (e.g., sparse vs. distributed encoding~\cite{morcos2018importance,zhou2018revisiting,bau17network,fong18net2vec}, invariance vs. sensitivity to geometric transformations~\cite{zeiler2014visualizing,lenc18understanding}),
or visualizing stimuli preferred by a network~\cite{zeiler2014visualizing,simonyan14deep,mahendran15understanding,nguyen2016synthesizing,bau2017network,olah2017feature,ulyanov18deep,mordvintsev2018differentiable}.
Another direction is to summarize a complex model with a simpler, more interpretable model (e.g., a sparse linear classifier or shallow decision tree)~\cite{bastani2017interpretability,lakkaraju2017interpretable,tan2018learning,zhang2018interpreting,zhang2019interpreting}.
Our work is most related to an approach introduced by several recent works~\cite{zhou2018interpretable,fong18net2vec,kim2018interpretability} that identify how semantic concepts are represented in a network by training linear probes on intermediate features to perform concept classification.
While these techniques focus on learning post-hoc how concepts are encoded, our method explicitly learns intermediate features that correspond to concepts.


\paragraph{Interpretability by design.}
In contrast to post-hoc approaches, ``interpretable-by-design'' paradigms focus on designing models that are explicitly interpretable. 
A number of works have proposed models that generate explanations alongside predictions.
A few papers utilize multiple modalities in their model to produce explanations~\cite{hendricks2016generating,zhang2017mdnet,huk2018multimodal}.
Another approach is to include an attention mechanism that constrains information flow~\cite{xiao2015application,lu2016hierarchical}.
Then, the attended features can be used as an explanation.
These are not explicitly designed to be human-interpretable, although~\cite{ross2017right} constrains attended features to match desired explanations.
A shortcoming of models optimized to produce explanations is that there is often a tradeoff between their explanatory and predictive components (e.g., a generated explanation may not be both faithful to the model and easily interpretable).

Another direction focuses on encouraging a model to have interpretable intermediate features.
Several works have introduced interpretable variational autoencoders~\cite{chen2016infogan,higgins2017beta} by encouraging the latent space to be disentangled (i.e., independent factors of variation).
Regarding image classifiers,~\cite{zhang2018interpretable} constrains features to be sparse, discriminative ``parts'' detectors, while~\cite{brendel2019approximating} introduces BagNets, interpretable classifiers that sum up evidence from small input patches.

Our work is most similar to~\cite{losch2019interpretability} and~\cite{marcos2019semantically}.
~\cite{losch2019interpretability} introduce ``semantic bottleneck networks,'' which encourage the features of the bottleneck between an encoder and decoder to align with semantic concepts.
Their design incurs a negligible loss in accuracy on the final segmentation task.
However, the relationship between the semantic bottleneck and output prediction is a highly non-linear decoder, making it difficult to study.
To improve the interpretability of the semantic bottleneck,~\cite{marcos2019semantically} use an semantic bottleneck based on attribute prediction and a linear layer to map onto the final task.
The linear mapping makes the relation between the concepts in the bottleneck and the final task easily interpretable; however, this forces each concept to contribute independently and linearly, with no regard for the presence of other concepts. 

\paragraph{Context.}
Visually similar attributes might be understood differently depending on what other elements are present, making contextualization important both for human and machine visual tasks~\cite{oliva2007role,barenholtz2014quantifying}.
\cite{lopez2017discovering} learns the causal relationship between pairs of object instances (e.g., cars and wheels) as well as the relationships between objects and background context, while~\cite{harradon2018causal} leverages a bayesian causal model to explore the impact of counterfactuals using concepts learned from self-supervision.
In order to leverage context, ScenarioNet~\cite{daniels2018scenarionet} proposes to find ``scenarios,'' groups of commonly co-occurring concepts, by learning a sparse dictionary on the co-occurrence matrix of concepts in the training set. 
These scenarios are then treated as classes and predicted jointly with the final task, allowing to see which scenarios are present in the image.
However, this does not necessarily mean that the final decision is conditioned on the detected scenarios.
We propose to connect the semantic bottleneck with the final output by using non-linear,  simple, and sparse relations, so that the mapping is transparent and easy to study, while being powerful enough to solve the visual task.
\section{Contextual Semantically Interpretable Bottleneck (CSIB)}

\begin{figure*}[!t]
    \centering
    \includegraphics[width=\textwidth]{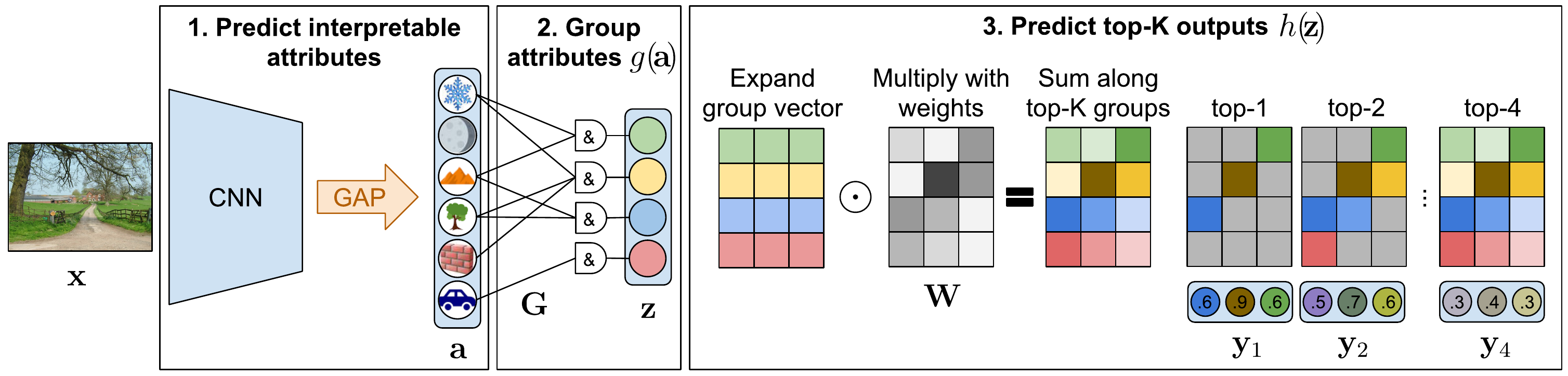}
    \caption{\textbf{Model overview.}
    Our CSIB model learns to 1., predict human-interpretable attributes, 2., form sparse groups of attributes that describe broader, task-relevant concepts, and
    3., output predictions using the top-$K$ groups.}
    \label{fig:flowchart}
\end{figure*}

Our proposed approach is illustrated in Fig.~\ref{fig:flowchart} and relies on two steps. First, we train a predictor of binary attributes using a standard \texttt{CNN}. The attributes are then computed by $\mathbf{a}=\texttt{CNN}(\mathbf{x})$ and all carry the semantics encoded by the attribute dataset. They can be object or more complex concepts contained in the images and multiple attributes can be present in a given image $\mathbf{x}$. Second, we train and interpretable function $f$ that uses the attributes to obtain a final prediction  $\mathbf{y}=f(\mathbf{a})=f(\texttt{CNN}(\mathbf{x}))$. 
The function $f$ should be simple enough so that a human observer can easily understand how each attribute contributes to the result. 
A common choice for a simple and interpretable function is a linear mapping $\mathbf{y}=\mathbf{W}\mathbf{a}$~\cite{marcos2019semantically}. However, such a function is unable to capture an important process for image understanding: \emph{contextualization}. This is because the contribution of an attribute to the output in the linear case is independent on the presence of other attributes.



Our solution consists of learning the groups of interaction by using a composition of two simple, but non-linear, functions: $\mathbf{z}=g(\mathbf{a};\mathbf{G})$, parametrized by the sparse matrix $\mathbf{G}$, which extracts relevant groups of attributes,  and $\mathbf{y}=h(\mathbf{z};\mathbf{W})$, parametrized by $\mathbf{W}$, which captures the relations between the groups and the output. 
In the following we detail the three elements that compose our model: the CNN that extracts attributes, and the functions $g$ and $h$. 

\subsection{Attribute prediction} We train a standard $\text{CNN}$ to predict the presence probability vector $\mathbf{a}\in[0,1]^A$ of \diego{}{$A$} attributes by minimizing the multi-label classification loss based on binary cross entropy, $\mathcal{L}_\text{attr}$, on an attribute dataset.
Training the model predicting the attributes typically needs labels not available for the dataset used for the final task $\mathbf{y}$. Therefore, to minimize $\mathcal{L}_{\text{attr}}$, we resort to an auxiliary dataset providing the attributes labels via image/attributes. Note that these two datasets can be disjoint and there is no need for images with both types of annotations. By choosing the appropriate set of attributes we are able to obtain a model that makes use of the desired visual cues while being invariant to undesired attributes.

\subsection{Attribute grouping function $g(\cdot)$} 
Given the attributes \textbf{a} predicted in the images, we now want to group them into semantically meaningful groups. To do so, we use a grouping function $\mathbf{z}=f(\mathbf{a})$ that groups attributes together into a vector of group presence probabilities $\mathbf{z}\in[0,1]^Z$\diego{}{, with $Z$ the number of groups}.
This function is parametrized by the sparse non-negative matrix $\mathbf{G}\in[0,1]^{Z\times A}$. Each row $\mathbf{G}_{i,:}$ represents one group and is constrained on the probability simplex ($G_{i,j}\geq 0, \sum_j G_{i,j}=1,\forall i$) by orthogonal projection after  each SGD
step~\cite{shalev2006efficient}. The output for group $z_i$ is computed as:
\begin{equation}
    z_i = \prod_{j=1\dots A} a_i^{\mathbf{G}_{i,j}}.
    \label{eq:softand}
\end{equation}
This corresponds to a weighted geometric mean and acts as a soft-AND logical function, which means that a group $i$ will only be fully active ($z_i=1$) if, for every attribute $j$ required by the group ($G_{i,j}>0$), the attribute is fully present ($a_j=1$).
Also, it suffices that one of these attributes is absent ($a_j=0$) to result in $z_i=0$. 
Since all the attributes for which $G_{i,j}>0$ must be present for the group to become active, $\mathbf{G}$ tends to become sparse during the learning process.  This is a direct consequence of the projection onto the simplex which is naturally sparse. 
To increase numerical stability, the operation is implemented as a standard linear mapping over the log probabilities of the attributes:
\begin{equation}
    \mathbf{z} = e^{\mathbf{G} \log(\mathbf{a}) },
    \label{eq:logsoftand}
\end{equation}
which allows us to use a numerically stable implementation of log-sum-exp.

\subsubsection{Unsupervised group pretraining}

The soft-AND function in Eq.~(\ref{eq:softand}) will output values close to zero if one or more of the attributes that correspond to a high weight $G_{i,j}$ are not present. Therefore, initializing $\mathbf{G}$ with random weights, encoding for random groups that are thus not very likely to exist, results in mostly inactive groups and a very low learning signal. To make sure that $\mathbf{G}$ is initialized with groups that are present in the dataset, we  first minimize the following loss on $\mathbf{Z}\in[0,1]^{B\times Z}$ corresponding to the concatenation for a batch of images with batch size $B$ :
\begin{equation}
   \mathcal{L}_\text{groups}(\mathbf{Z})=
    \mathcal{L}_\text{on}(\mathbf{Z})+
    \mathcal{L}_\text{off}(\mathbf{Z})+
    \mathcal{L}_H(\mathbf{Z}).
\end{equation}
The  first two terms are designed to encourage the groups to become diverse. For a group to be of any use, it needs to be active in at least a few images. In addition, we want to make sure that at least one group is active per image. For this reason, we encourage the highest values along each row and each column of $\mathbf{Z}$ to be close to one, the highest possible value:
\begin{equation}
    \mathcal{L}_\text{on}(\mathbf{Z})
    =-\sum_{i={1}}^{Z}\max_u (Z_{ui})- \sum_{u={1}}^{B}\max_i(Z_{ui}).
\end{equation}
At the same time, we want to make sure that no particular group is active in all the samples of the batch, because such group would not be a discriminative one. We therefore minimize the maximum of the lowest per-group values:
\begin{equation}
   \mathcal{L}_\text{off}(\mathbf{Z})
    = \max_u(\min_i({Z}_{u,i})).
\end{equation}
However, this is not enough to guarantee the diversity in the groups. Ideally, we would want the batch-wise vector of group activations $\mathbf{Z}_{:,i}$ and $\mathbf{Z}_{:,j}$ of any two groups to be as different as possible. Simultaneously, we would like the groups to help discriminate between images, and thus the sample-wise vectors of group activations $\mathbf{Z}_{u,:}$ and $\mathbf{Z}_{v,:}$ of any pair of images should also be as different as possible. We encourage this by maximizing the cross-entropy $H(\mathbf{u},\mathbf{v}) = -\sum_i u_i\log(v_i)$ between all pairs of per-group activation vectors and all pairs of per-sample activation vectors: 
\begin{equation}
    \mathcal{L}_{H}(\mathbf{Z})
    = 
    -\sum_{i,j\neq i} H\left(\frac{\mathbf{Z}_{:,i}}{\sum_k Z_{k,i}},\frac{\mathbf{Z}_{:,j}}{\sum_k Z_{k,j}}\right)
    -
    \sum_{i,j\neq i} H\left(\frac{\mathbf{Z}_{i,:}}{\sum_k {Z}_{i,k}},\frac{\mathbf{Z}_{j,:}}{\sum_k {Z}_{j,k}}\right)
\end{equation}
Note that the regularization term above will have the effect of promoting the activations across groups and images to the maximally independent, tending towards source separation. 
Minimizing $\mathcal{L}_\text{groups}$ provides $\mathbf{G}$ with a set of initial groups that do occur in some images ($\mathcal{L}_\text{on}$) but not in all ($\mathcal{L}_\text{off}$) and that are discriminative  and different from each other ($\mathcal{L}_H$).

\subsection{Output contribution function $h(\cdot)$}

Given the group activations $\mathbf{z}$, we want a function
$\mathbf{y}=h(\mathbf{z})$ that produces the desired final output $\mathbf{y}\in\mathbb{R}^Y$. Function $h$ is parametrized by matrix $\mathbf{W}\in\mathbb{R}^{Y\times Z}$. 

We want as few groups as possible to contribute to the output $\mathbf{y}$. This can be enforced by taking only the top-$K$ most contributing groups to compute $\mathbf{y}$, as proposed in~\cite{DKA2019}. The following steps have to be taken:
\begin{itemize}
    \item A matrix element-wise multiplication $\mathbf{Y}=\mathbf{W}\circ\mathbf{z}$, where $\mathbf{z}$ is broadcasted to the shape of $\mathbf{W}\in\mathbb{R}^{Y\times Z}$.
    \item A sparsification of $\mathbf{Y}$ by keeping only the top-$K$ values in each row and setting the rest to zero.
    \item Row-wise sum to obtain $\mathbf{y}$.
\end{itemize}
    
In order to avoid choosing a value of $K$ a priori, we compute $\mathbf{y}$ using multiple $K$ values and apply a loss to each output. The specific loss used at this stage is problem-dependent (\emph{e.g.} MSE for regression, cross entropy for classification, \emph{etc.}). The average of such losses is the final output loss, $\mathcal{L}_y$.

\subsection{CSIB training strategy}\label{ssec:train}

The training procedure to minimize the described losses consists of three steps:
\begin{enumerate}
    \item Train the CNN to predict the concepts in the semantic bottleneck by minimizing $\mathcal{L}_{\text{attr}}$. Note that this provides no learning signal to $\mathbf{G}$ nor $\mathbf{W}$. 
    \item Keeping the weights of the CNN frozen, minimize $\mathcal{L}_{\text{groups}}$ to initialize $\mathbf{G}$ with relevant and discriminative groups. 
    \item Finetune the whole model end-to-end on the final task by minimizing $\mathcal{L}_{\text{attr}} + \lambda\mathcal{L}_y$, with $\lambda<<1$ to ensure that the performance on attribute prediction is not degraded.
\end{enumerate}

\section{Experiments in landscape scenicness prediction}


\subsection{Experimental set-up}

In the experiments below, we aim at predicting the  scenicness (\emph{i.e.} landscape beauty) score of a collection of images.
The training images come from the ScenicOrNot~\cite{scenicornot} dataset, collected across Great Britain, and where each image has an average scenicness score (between 1 and 10), obtained by crowdsourcing.
Out of the 212,104 available images, we used the first 180,000, ordered by image ID, for training, the following 5,000 for validation and the rest we held out for testing.
Given the subjectivity of the final task, we want to make the reasoning of the model explicit by using a semantic bottleneck that detects attributes occurring in the image as an intermediate task: therefore,
the semantic bottleneck is trained to predict the presence of the 102 classes of the SUN Attributes~\cite{patterson2014sun} dataset.
We use the same train-test splits as in~\cite{patterson2014sun}. \diego{}{Previous works have already established that there is a correlation between some of these attributes and scenicness~\cite{seresinhe2017using,workman2017understanding}, and with SCIB we aim at constraining this further and explain scenicness using exclusively this pre-defined set of attributes}.


For attribute prediction, we finetune a ResNet-50~\cite{he2016deep}, pre-trained on ImageNet~\cite{ILSVRC15}. We remove the last layer of the pre-trained model and add a $1\times 1$ convolutional layer to map down the 2048 activation maps to the 102 SUN attributes, followed by a global average pooling. The model is trained using Stochastic Gradient Descent (SGD) with 0.9 momentum for 20,000 iterations with a batch size of 10.
The learning rate is initially 0.002 and is decayed by a factor of 4 after 10,000 and 15,000 iterations.
In the second step (group initialization), we initialize the sparse grouping matrix $\mathbf{G}$ with 150 groups in an unsupervised way by minimizing $\mathcal{L}_\text{groups}$.
The learning rate was fixed at 0.002 for 4,000 iterations with a batch size of 100.
Note that, as mentioned in section~\ref{ssec:train} above, the ResNet-50 base model was left frozen, allowing for a larger batch size, which is important to capture the diversity between groups in $\mathcal{L}_\text{groups}$.
As a last step (finetuning), $\lambda\mathcal{L}_y$ is minimized along with $\mathcal{L}_\text{attr}$ for 50,000 iterations and $\lambda=0.1$.
This time the whole model is trained end-to-end and two batches of size 10 are used in every iteration, one from SenicOrNot and one from SUN Attributes.
The learning rate is initially 0.002 and is decayed by a factor of 4 after 10,000 and 20,000 and 30,000 iterations. We train simultaneously with nine levels of top-$K$ sparsity: $K={1,2,...,8,150}$. Since $\mathbf{W}$ is initialized will all zeros, the dense branch is important to make sure that all groups receive a learning signal.
The bias of the last layer ($\mathbf{W}$) is fixed to the average scenicness value on the training set, a score of 4.43, and is kept constant.


\subsection{Numerical comparisons within ScenicOrNot}\label{ssec:resquant}

\begin{figure}[!t]
\centering
\floatbox[{\capbeside\thisfloatsetup{capbesideposition={right,top},capbesidewidth=0.5\textwidth}}]{figure}[\FBwidth]
{
\caption{Average number of attributes with a nonzero contribution to the final output against the resulting Kendall's $\tau$ score. The number of nonzeros was varied by increasingly pruning the smaller contributions. Our model allows for much more concise explanations.}
\label{fig:nonzeros}}
{\includegraphics[width=0.45\textwidth]{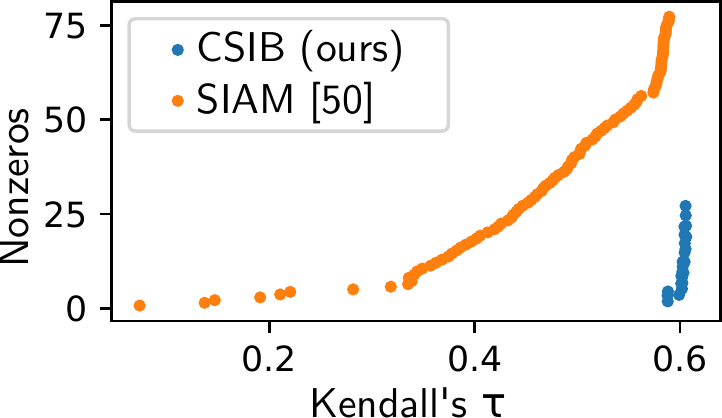}}
\end{figure}

\begin{figure}[!t]
    \centering
   \includegraphics[width=\columnwidth]{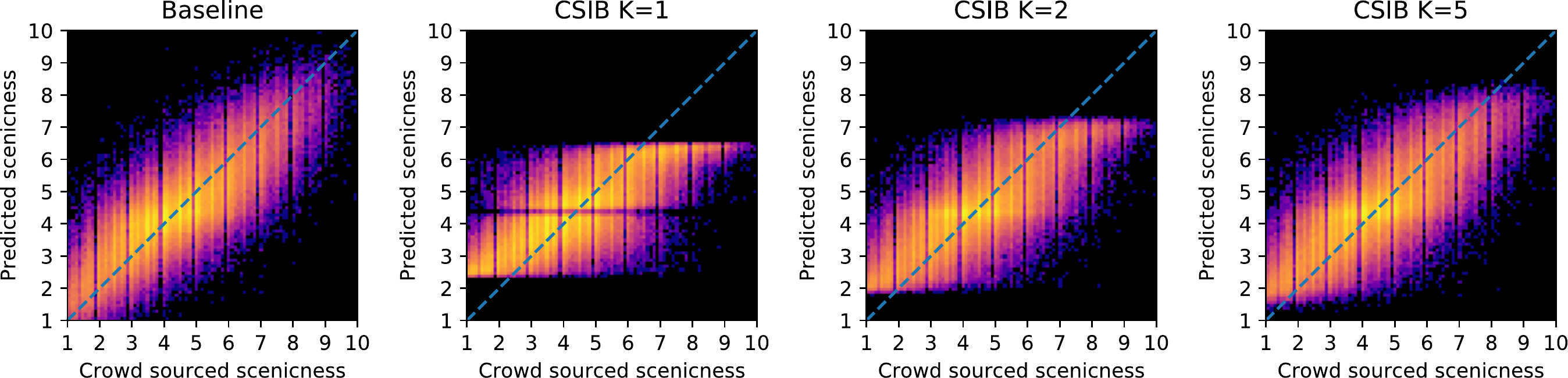}
    \caption{Scatter plots showing our CSIB model's predictions for different $K$, which controls the number of groups used to make the final prediction (see~\cref{fig:sankey_model_son} for groups).}
    \label{fig:scatter}
\end{figure}
\begin{table}[b!]
	\centering
	\begin{tabular}{r|c|c|c|c|c|}
		\cline{3-6}
		 \multicolumn{2}{c|}{} & \multicolumn{4}{|c|}{CSIB}\\
		\cline{2-6}
		& baseline &  $K=1$ &  $K=2$ &  $K=5$ &  $K=7$\\ \hline
		\multicolumn{1}{|l|}{SoN Kendall's $\tau$ } & 0.645 & 0.580 & 0.603 & 0.609 & 0.609\\ \hline
		\multicolumn{1}{|l|}{SoN RMSE} & 0.940 & 1.111 & 1.037 & 1.018 & 1.019\\ \hline\hline
	    \multicolumn{1}{|l|}{SUN AP} & 0.610 & \multicolumn{4}{|c|}{0.601}\\\hline
	\end{tabular}
\caption{\textbf{Task performance.} ScenicOrNot (SoN) results are reported using Kendall's $\tau$ ranking metric and root mean square error (RMSE); average precision (AP) is reported for SUN (higher is better for $\tau$ and AP; lower is better for RMSE). Performance plateaus at $K = 5$ for our CSIB model; our model underperforms the baseline.}
\label{tab:son}
\end{table}

\begin{figure}[!t]
    \centering
    \includegraphics[width=\columnwidth]{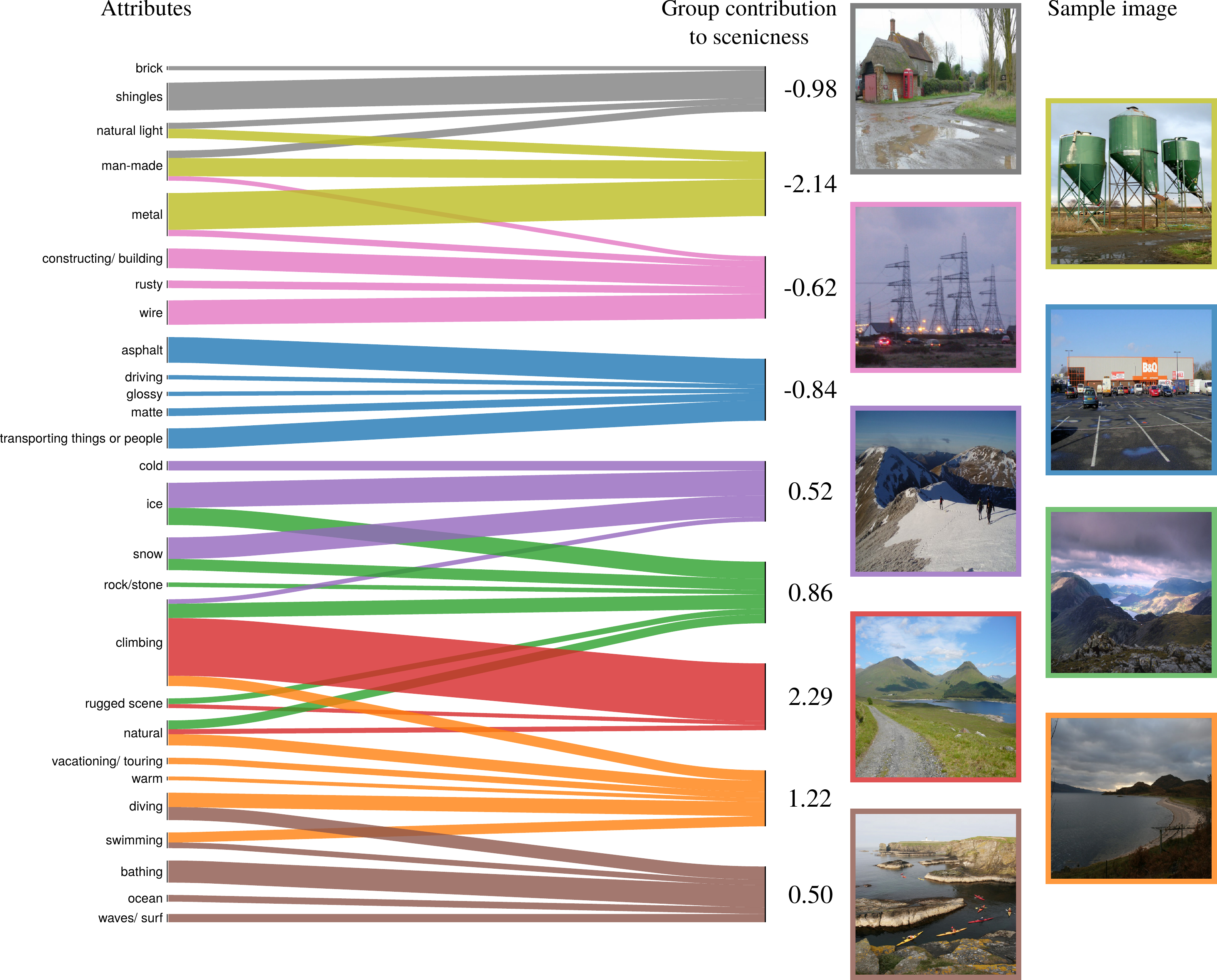}
    \caption{\textbf{Learned groups for scenicness estimation.}
    Line thickness denotes the contribution of a SUN Attribute (left) to a group. For each group, we show its scenicness score (middle) and an example image (right).
    The groups and their scores appear coherent and consistent with the task.
    They are also sparse and interpretable: only 27 of 102 attributes are chosen and only 9 groups (out of a possible 150) become relevant.}
    \label{fig:sankey_model_son}
\end{figure}

On the ScenicOrNot test set, both CSIB and the baseline are able to generalize  well, with CSIB showing a small drop in performance in terms of Kendall's $\tau$~\cite{kendall1938new} and root mean square error (RMSE), comparable to the one observed in~\cite{marcos2019semantically} (see Tab.~\ref{tab:son}). \diego{}{However, \cite{marcos2019semantically} requires many more attributes to contribute to the result compared to CSIB, as shown in Fig.~\ref{fig:nonzeros}, making the explanations provided by CSIB more desirable in terms of the number of required \emph{cognitive chunks}~\cite{doshi2017towards}. This highlights the effectiveness in terms of sparsification of the proposed constrained optimization.}
\diego{We observe a smaller loss in performance on attribute prediction, measured as the average precision against the union of all the annotators in SUN attributes}{In addition, we observe only a minor degradation in the performance on the task of attribute prediction with respect to a baseline trained exclusively on that task}. 
CSIB enables us to choose at test time 
the number of groups that can contribute to the final result by setting the $K$ parameter in the top-$K$ activation layer. Fig.~\ref{fig:scatter} depicts the results for $K \in \{1,2,5\}$ in the form of scatter plots.
When using $K=1$, CSIB is not able to predict very high or very low values, since the maximum deviation from the average it can predict is $[-2.14,2.29]$, which corresponds to the contribution of the most contributing groups (see red and yellow groups in Fig.~\ref{fig:sankey_model_son}). At the same time, values close to the average are also missed, since the top-1 layer is required to choose the single most contributing group among the groups present, forcing the output away from the average. This undesirable behaviour is already corrected by setting $K=2$. The accuracy saturates when using $K=5$, where the model is capable of predicting more extreme values of scenicness in a comparable way to the baseline, although it maintains a bias towards the average in the extreme cases. Such a small value of $K$, together with the sparsity of $\mathbf{G}$, allows to easily understand the relations encoded in CSIB (see Section~\ref{ssec:resqual}). 
\diego{}{We observed that finetuning the whole model end-to-end (step 3 in Section~\ref{ssec:train}) was important to obtain the mentioned results, with a Kendall's $\tau$ of 0.468 before finetuning.}

\subsection{Visualization of the model}\label{ssec:resqual}

\paragraph{Entire model.} The semantics captured by SCIB, together with the high sparsity, allow us to comprehend the reasoning used to compute the output from the attributes in the semantic bottleneck. Fig.~\ref{fig:sankey_model_son} depicts the model by showing the contribution of each attribute to the groups (the weights in $\mathbf{G}$) that contribute more than 0.5 score points towards the scenicness values. This already provides a good understanding of the relations learned and encoded in the model. For instance, the last two groups (orange and brown) show that ``diving'' and ``swimming'' are assigned higher scores if they co-occur with ``climbing'' and ``natural.'' We also see that it typically assigns high scores to wilderness-related attributes and low scores for those related to man-made elements. In the same figure, we also provide an image that scores strongly for that group.

\begin{figure}[!t]
    \centering
    \includegraphics[width=\columnwidth]{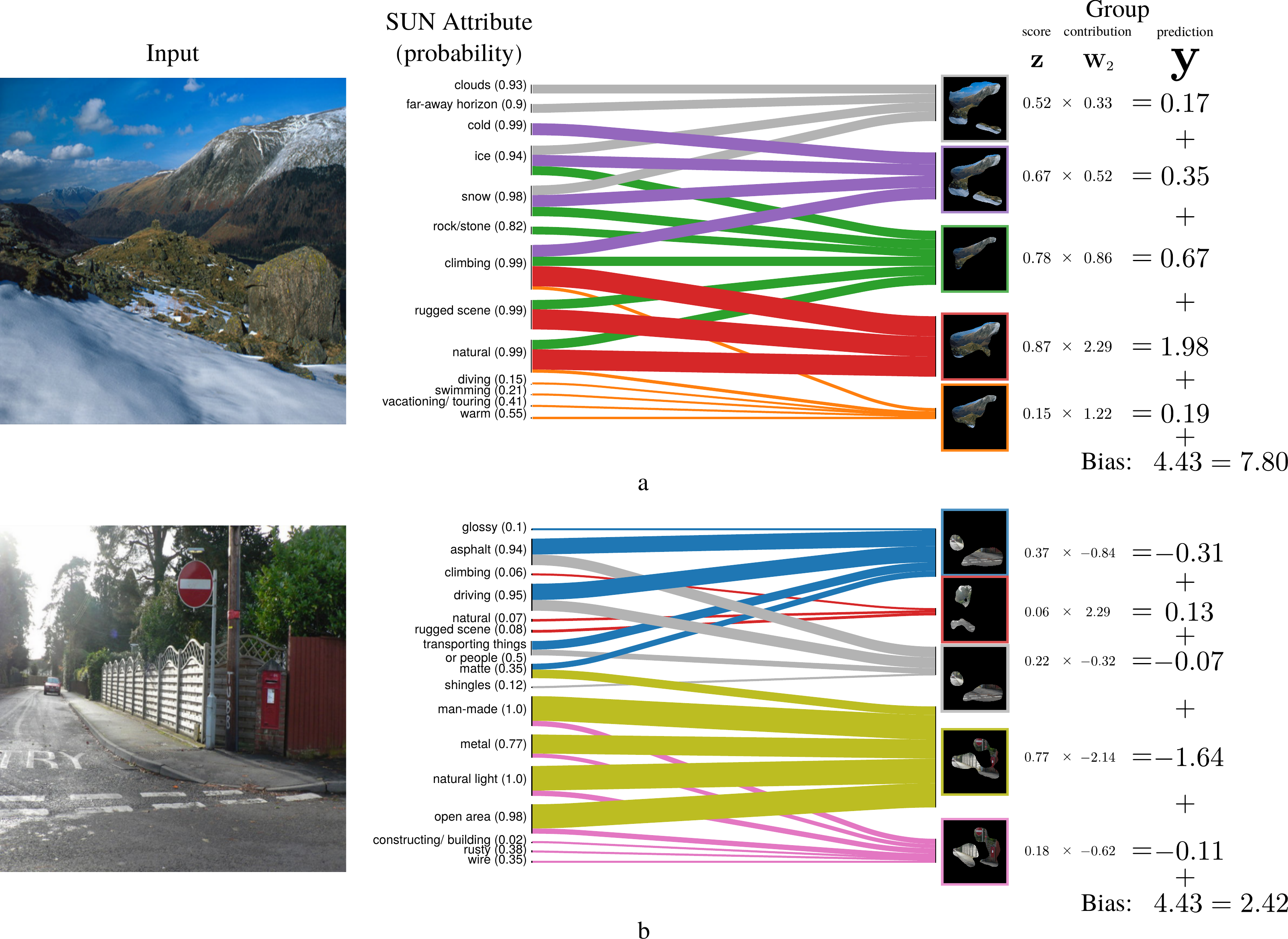}
    \caption{Diagram of attribute contribution for a scenic image (top: GT = 9.43, baseline = 8.39) and an unscenic one (bottom: GT = 1.6, baseline = 2.18). These explanations point to a sensible decision-making process.}
    \label{fig:sankey_indiv_2}
\end{figure}

\begin{figure}[!t]
    \centering
    \includegraphics[width=\columnwidth]{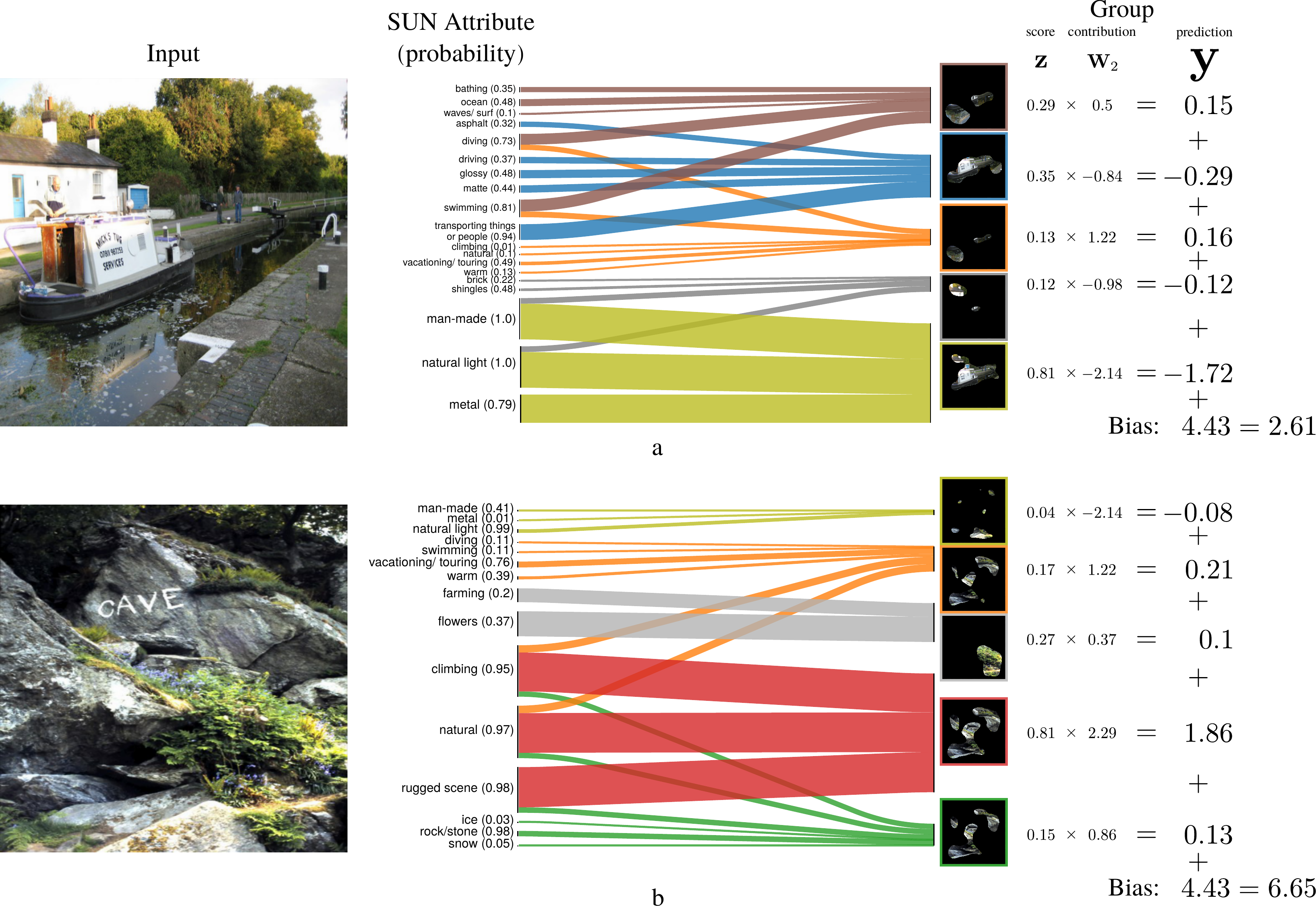}
    \caption{Diagram of attribute contribution for an underpredicted image (top: GT = 7.17, baseline = 3.66) and an overpredicted one (bottom: GT = 2.11, baseline = 6.65). Our CSIB model allows us to understand why it disagrees with ground truth annotations. \vspace{-0.3cm}}
    \label{fig:sankey_indiv_1}
\end{figure}

\paragraph{Individual results.}
Individual decisions for specific images are also easily interpretable. Using the activations in CSIB, we can now visualize which paths are followed to reach the final decision. Fig.~\ref{fig:sankey_indiv_2} shows two examples with $K=5$. In this figure, the thickness of the lines is proportional to the contribution of the attribute to the group, which depends on the presence of the other attributes required by the group due to the multiplicative nature of Eq.~(\ref{eq:softand}). The part of the images contributing the most to the groups is depicted using thresholded activation maps. In these two cases, we can see how the explanations suit the images and our preconceptions of landscape beauty, with the first one rated with a 7.8/10 due to the rugged snowy mountain scene and the second one a mere 2.4/10 because of its man-made nature. On the other hand, Fig.~\ref{fig:sankey_indiv_1} depicts the same visualization for two images in which there is a strong disagreement with the crowdsourced value. In the first case, the man-made look of the image and the transport-related aspect of the boat trigger the model to predict a low score, while in the second case the ruggedness and climbing-related aspects dominate, while the graffiti and the narrow view of the image are ignored, since these cannot be captured by the  attributes used.

\subsection{Validation of the group predictions by geographical distribution}\label{ssec:geo}

\begin{figure}[!t]
    \centering
        \includegraphics[width=\columnwidth]{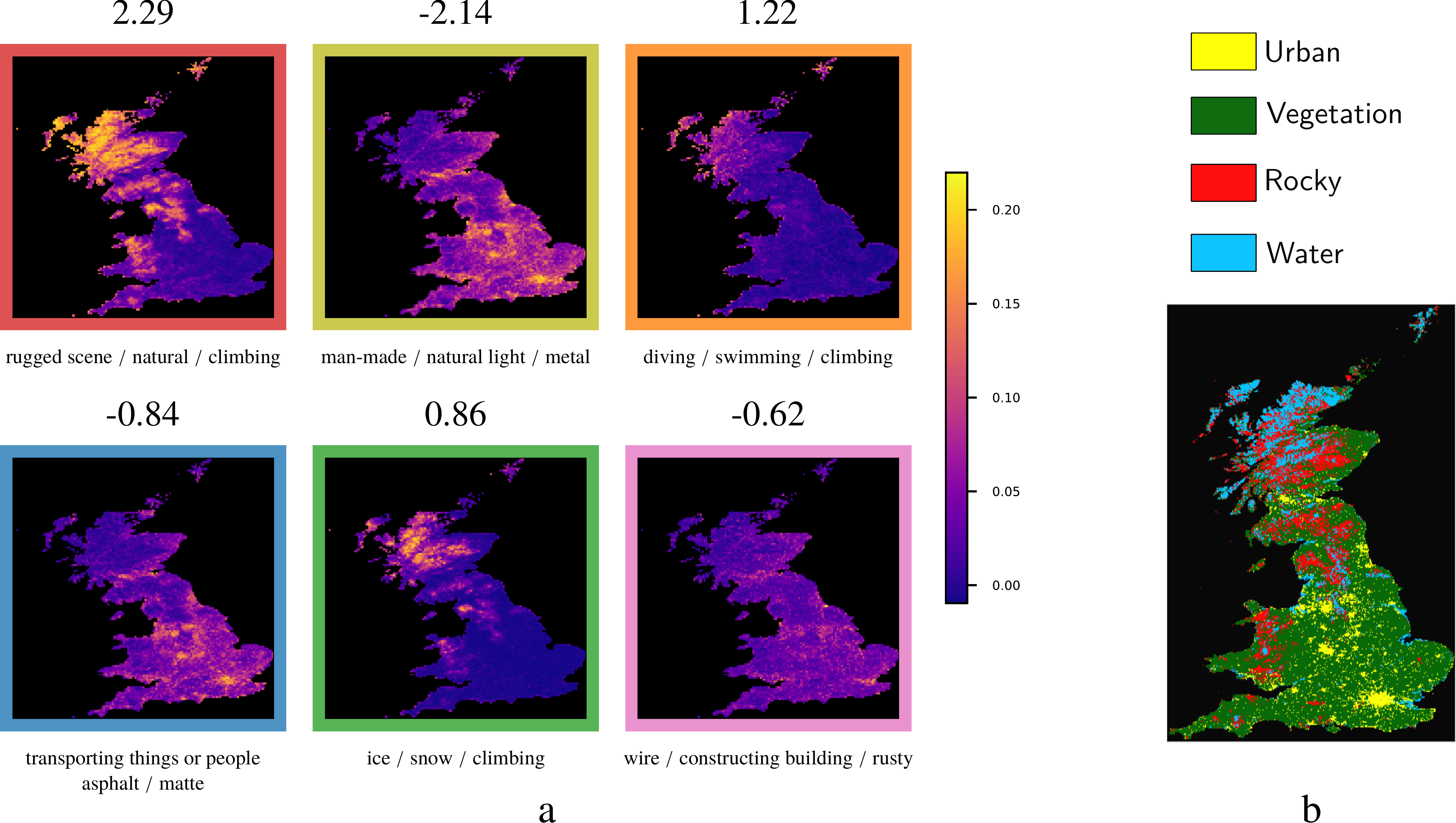}
    
    \caption{
    \textbf{a. Geographical distribution of group activations.}
    For each image, the number on top is the group's scenicness score, and the three most contributing attributes for the group are shown below.
    Group colors are taken  from~\cref{fig:sankey_model_son}.
    \textbf{b. Landcover of Great Britain.} Data from~\cite{corine}. \vspace{-0.2cm}}
    \label{fig:maps}
\end{figure}

Being the attributes learned from a dataset that is disjoint from the one used for the final task (\emph{i.e.} we have no test set of the 102 SUN attributes on the ScenicOrNot images), we evaluate the performance of the attribute prediction on the SoN images qualitatively, by mapping the geographical distribution of the average activation of the  groups (Fig.~\ref{fig:maps}a) and comparing them to the 2012 CORINE land-cover map~\cite{corine} of Great Britain (Fig.~\ref{fig:maps}b). The learned groups with mountain-related attributes show a good overlap with the bare soil and rock surfaces in the landcover map, and the group that also includes snow and ice is more present in mountainous regions of Scotland. The groups with man-made attributes overlap with urban areas, and the ones with water activities are most active along the coast and in the lake filled northwest. These results suggest a good performance of the attribute detector on the SoN image dataset.

\subsection{Generalization: numerical results on 1.7M Flickr images}\label{ssec:flickr}

\begin{figure}
    \centering
    \includegraphics[width=\columnwidth]{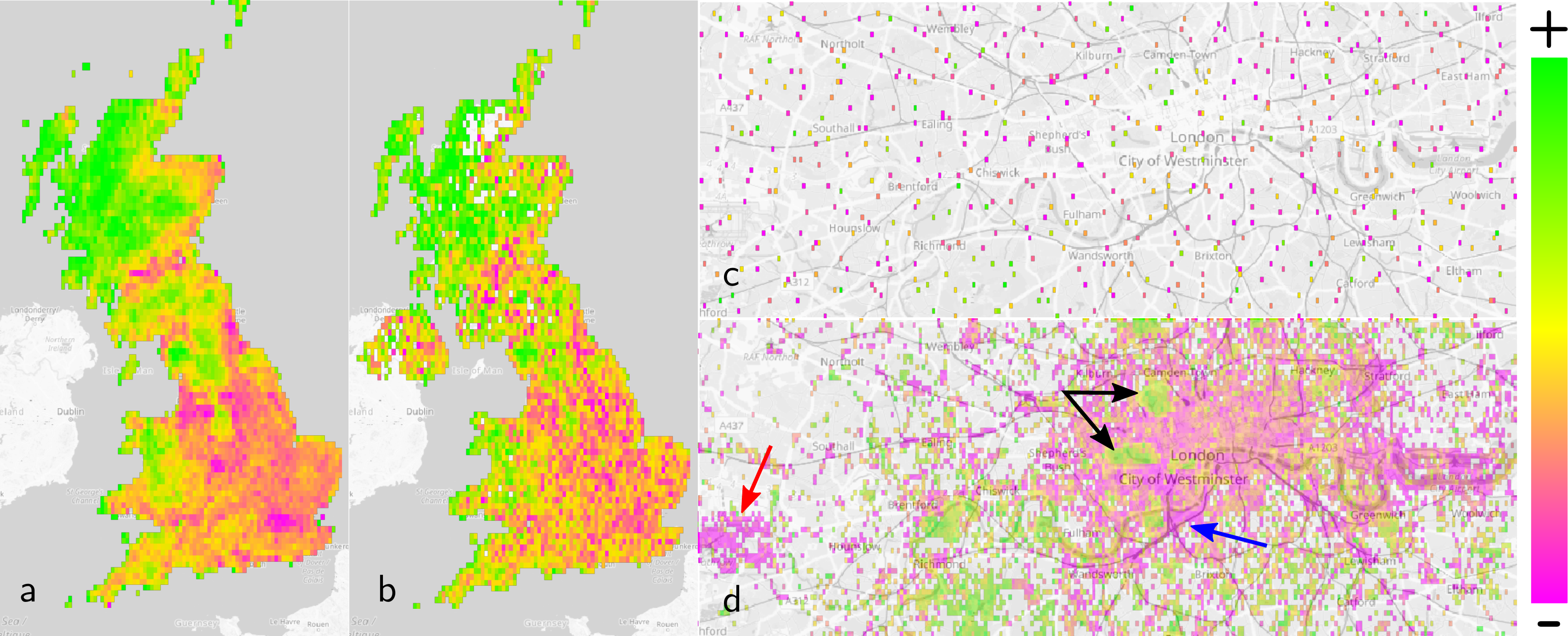}
    \caption{
    Geographical distribution of scenicness at the national level (a and b) and in London (c and d).
    Panels a and c show the ground truth scenicness interpolated from ScenicOrNot; panels b and d show our CSIB model's predictions.
    In panel d, a few London regions are quite salient, such as Hyde and Regent's parks (black arrows), Heathrow airport (red arrow) and a railway intersection (blue arrow). \vspace{-0.1cm}} 
    \label{fig:flickr_maps}
\end{figure}

In order to test its generalization capabilities, we applied both CSIB and the baseline over a large set of 1.7 million geo-located outdoor images obtained from Flickr. Although no scenicness ground truth is available for these images, we can create a map of scenicness based on the values predicted on the Flickr dataset (depicted in Fig.~\ref{fig:flickr_maps}b, c, e and f), and compare it to the map obtained using the ScenicOrNot ground truth (Fig.~\ref{fig:flickr_maps}a and d). 
In Tab.~\ref{tab:flickr} we show the results of comparing these values averaged over grids of different size ($5000\times 5000$, $500\times 500$ and $50\times 50$ bins across the region), and we consistently see that the results of the baseline and CSIB are numerically equivalent, with the CSIB being slightly better in terms of RMSE and behind in terms Kendall's $\tau$. This suggests that the better performance of the baseline on ScenicOrNot might be partially attributable to over-fitting to the dataset, and that restricting the model to be inherently interpretable with CSIB reduces its capacity to overfit, but not the capability to generalize. Fig.~\ref{fig:flickr_maps} shows maps of scenicness at two different scales (100 and 5000 bins) using the ScenicOrNot ground truth and CSIB results on the Flickr images. Fig.~\ref{fig:flickr_maps}a/b at the country level, showcases the agreement of both maps. Fig.~\ref{fig:flickr_maps}c/d show the maps for London. At this scale the sparsity of ScenicOrNot becomes apparent. On the map produced by CSIB, scenicness seems to be predicted highest in green areas (such as the Hyde and Regent's parks) and lowest in areas of transport infrastructure, such as railroads and airport (see Fig.~\ref{fig:flickr_maps}d). These conclusions regarding the notion of scenicness validate the relations captured by CSIB from the dataset and that are clearly visible and interpretable from the model itself (as seen in Fig.~\ref{fig:sankey_model_son}). \vspace{-0.2cm}

\begin{table}[h!]
	\centering
	\begin{tabular}{r|c|c|c|c|c|c|}
		\cline{2-7}
		 & \multicolumn{2}{|c|}{5000 bins} & \multicolumn{2}{|c|}{500 bins} & \multicolumn{2}{|c|}{50 bins}\\
		\cline{2-7}
		& baseline & CSIB & baseline & CSIB & baseline & CSIB\\ \hline
		\multicolumn{1}{|l|}{Kendall's $\tau$ } & 0.399 & 0.391 & 0.384 & 0.382 & 0.624  & 0.621 \\ \hline
		\multicolumn{1}{|l|}{RMSE} & 1.528 & 1.497 & 1.213 & 1.166 & 0.749 & 0.679 \\  \hline
	\end{tabular}
\caption{\textbf{Performance on Flickr images.}
We evaluate models on 1.7M Flickr images in Great Britain and bin the predictions spatially at different scales; we then compare the spatial predictions against ScenicOrNot ground truth averages.
Our CSIB model performs comparably to the baseline (higher is better for $\tau$; lower is better for RMSE). \vspace{-0.5cm}}
\label{tab:flickr}
\end{table}

\section{Conclusion}
We presented a paradigm to make the decision making process of a CNN inherently interpretable, which we call a Contextual Semantic Interpretable Bottleneck (CSIB).
A standard CNN is trained to detect human-interpretable attributes.
These attributes are then used to determine the final decision in a contextual and sparse manner (\emph{i.e.} the meaning an attribute takes on is dependent on what other attributes are present, and only a select subset of attributes are used to form a small number of groups).
This makes it possible to understand what relationships the model has learned by simply inspecting its weights as well as which of these relationships have been used for an individual image.
Note that CSIB requires an auxiliary dataset containing attributes relevant to the final task.
Nevertheless, the same attribute predictor can be reused for multiple downstream tasks; this would also enable model comparisons \emph{across} tasks and reveal what attribute groupings are relevant to which tasks.

We demonstrate the validity of our method on a scenicness estimation task; we use the ScenicOrNot dataset and also evaluate on a large set (1.7M images) of real-world images from Flickr.
CSIB is able to generate a map of scenicness to the same level of accuracy as that of a non-interpretable baseline.
Lastly, we show how visualization techniques can be combined in order to explain what (and how) visual information has been leveraged in our model's decision making process.
This allows us to understand the instances in which our model disagrees with the labelled annotation, among other things.
In conclusion, we introduce a novel architecture that is both inherently interpretable and powerful enough so as to not sacrifice in real-world performance.
This suggests that the assumed tradeoff between interpretability and performance may not always be necessary.
\bibliographystyle{splncs}
\bibliography{egbib}
\end{document}


\pagestyle{headings}
\mainmatter
\def\ECCVSubNumber{3278}  

\title{Contextual Semantic Interpretability} 

\titlerunning{ECCV-20 submission ID \ECCVSubNumber} 
\authorrunning{ECCV-20 submission ID \ECCVSubNumber} 
\author{Anonymous ECCV submission}
\institute{Paper ID \ECCVSubNumber}

\bibliographystyle{splncs04}

\section*{Supplementary material for paper ID 959}

\appendix

\section{Sensitivity to top-$K$}

\vspace{-0.1cm}

\begin{figure}
    \centering
    \includegraphics[width=0.49\columnwidth]{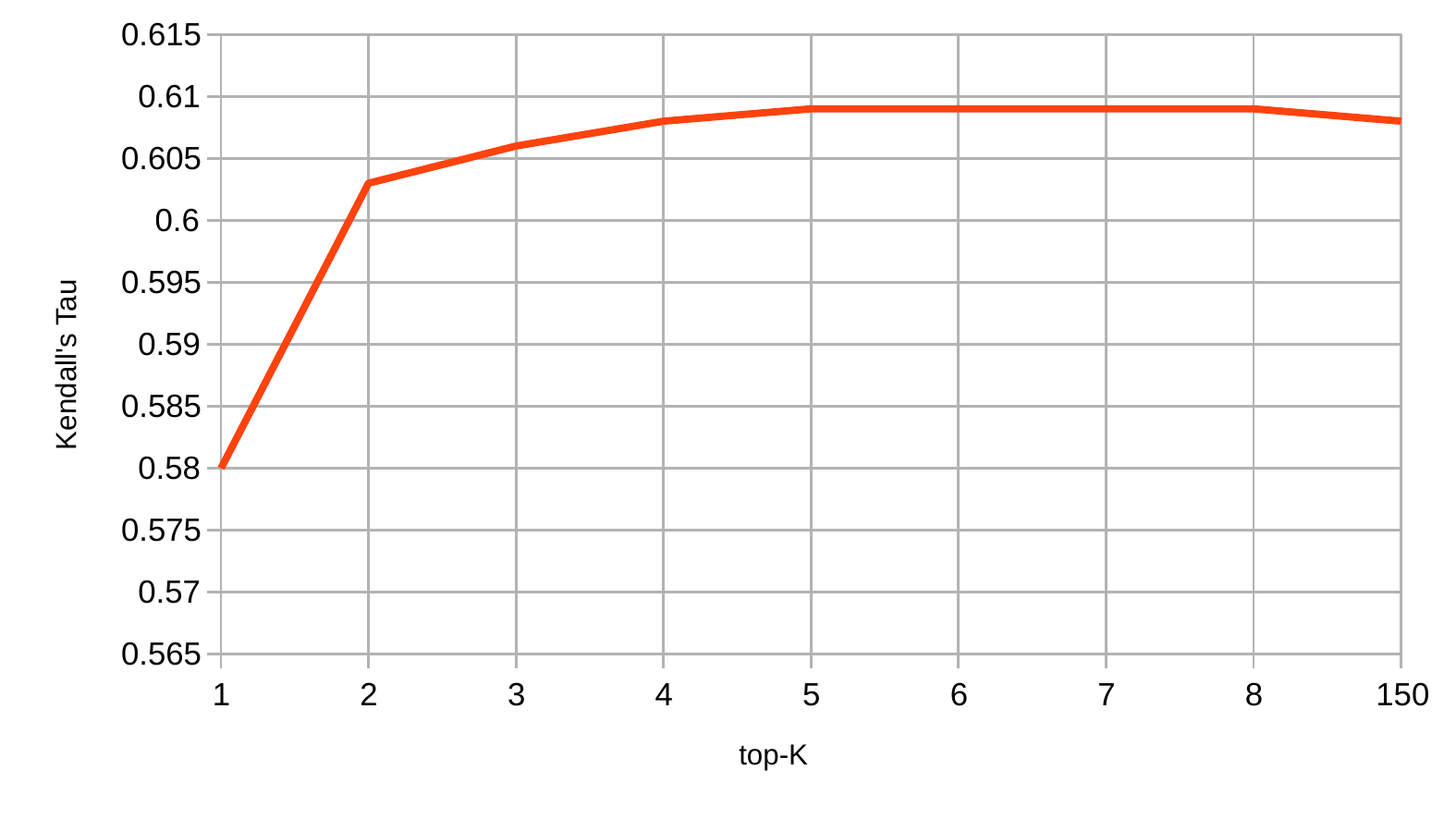}
    \includegraphics[width=0.49\columnwidth]{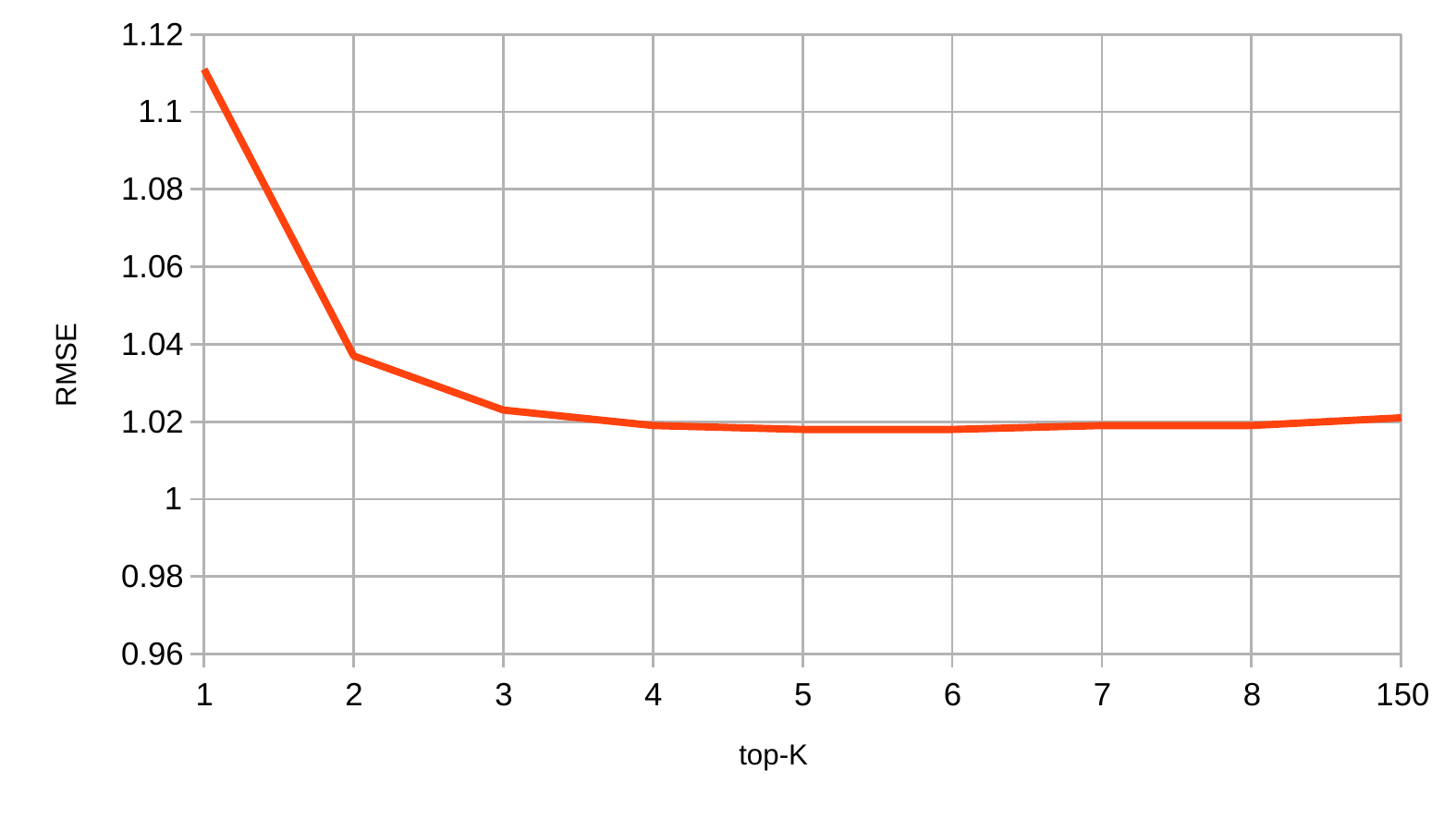}
    \caption{Performance for different values of $K$ in the top-$K$ operation, in terms of Kendall's $\tau$ (left) and RMSE (right). A top-$K$ of 150 represents the dense model. }
    \label{fig:topk}
\end{figure}

Fig.~\ref{fig:topk} shows the relationship between CSIB model performance and $K$ sparsity. We can see how the performance saturates for values higher than $K$=5.

\vspace{-0.1cm}

\vspace{-0.3cm}

\section{Map interpretability over London}

\vspace{-0.1cm}

\begin{figure}
    \centering
    \includegraphics[width=0.45\columnwidth]{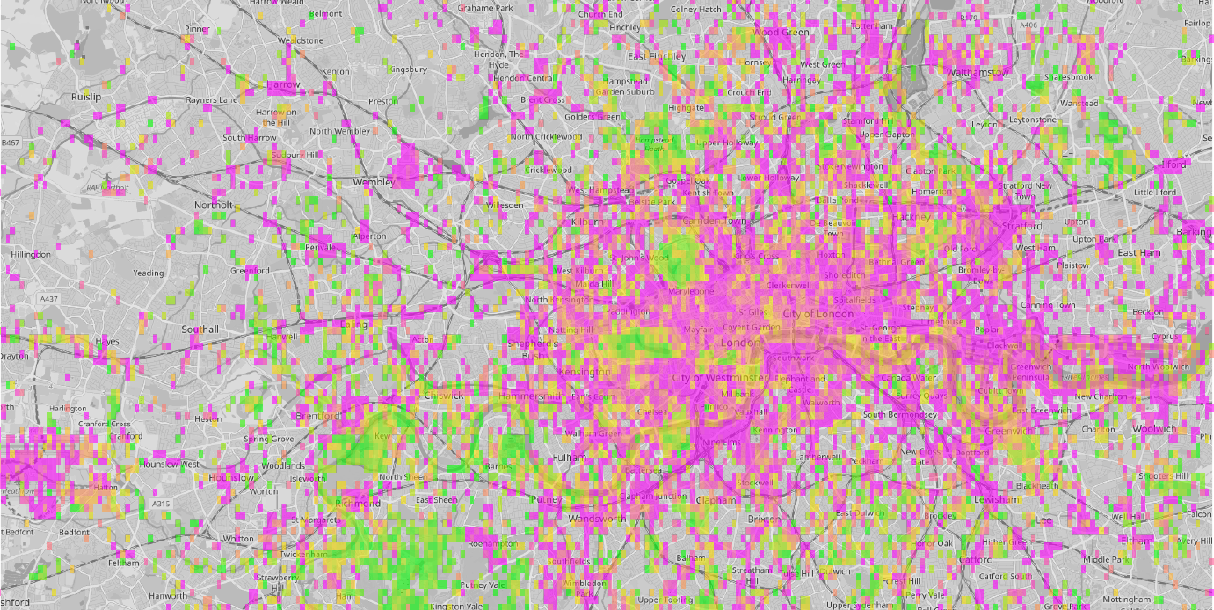}
    \includegraphics[width=0.45\columnwidth]{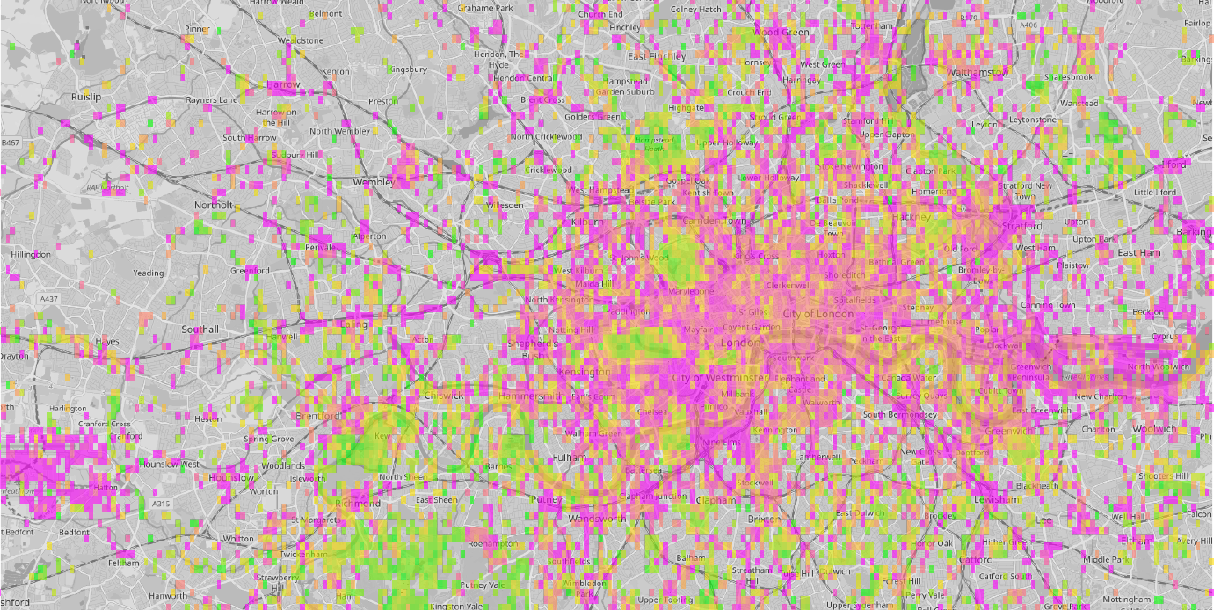}
    \includegraphics[width=0.02105\columnwidth]{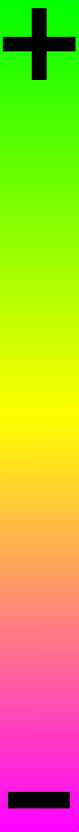}
    \caption{Map of scenicness over London from the Flickr images obtained using the baseline ResNet-50 (left) and the proposed interpretable CSIB (right).}
    \label{fig:scenicness}
\end{figure}

\begin{figure}
    \centering
    \begin{tabular}{c c}
      \includegraphics[width=0.45\columnwidth]{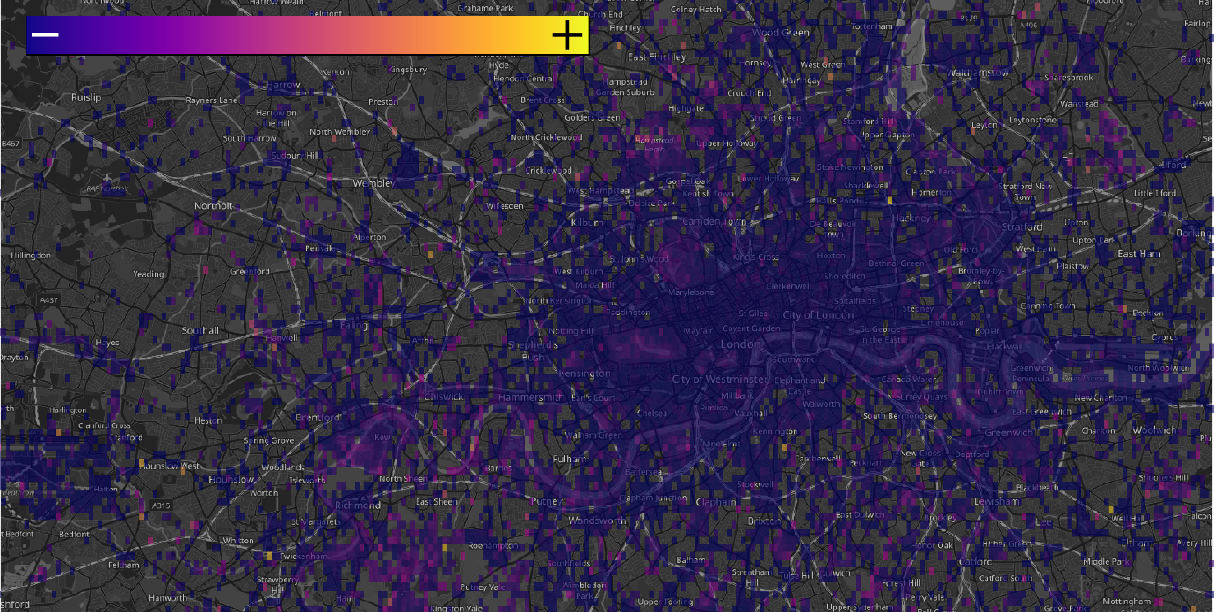}   &  \includegraphics[width=0.45\columnwidth]{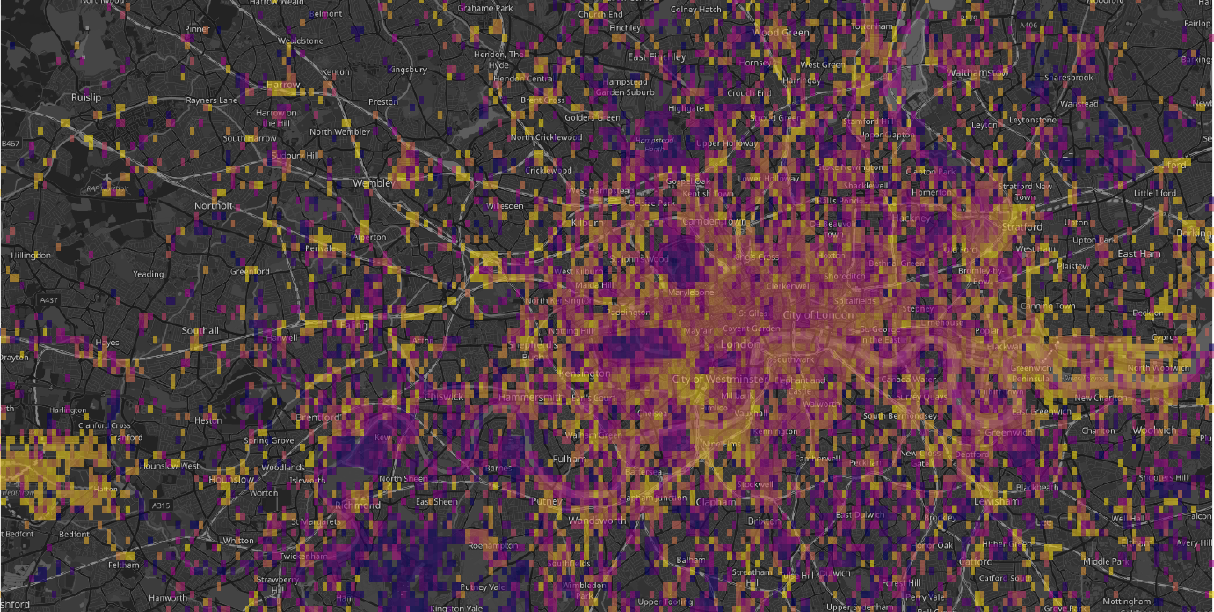}  \\
       (a) climbing, natural, rugged scene (2.29)  & (b) metal, natural light, man-made (-2.14)  \\
       \includegraphics[width=0.45\columnwidth]{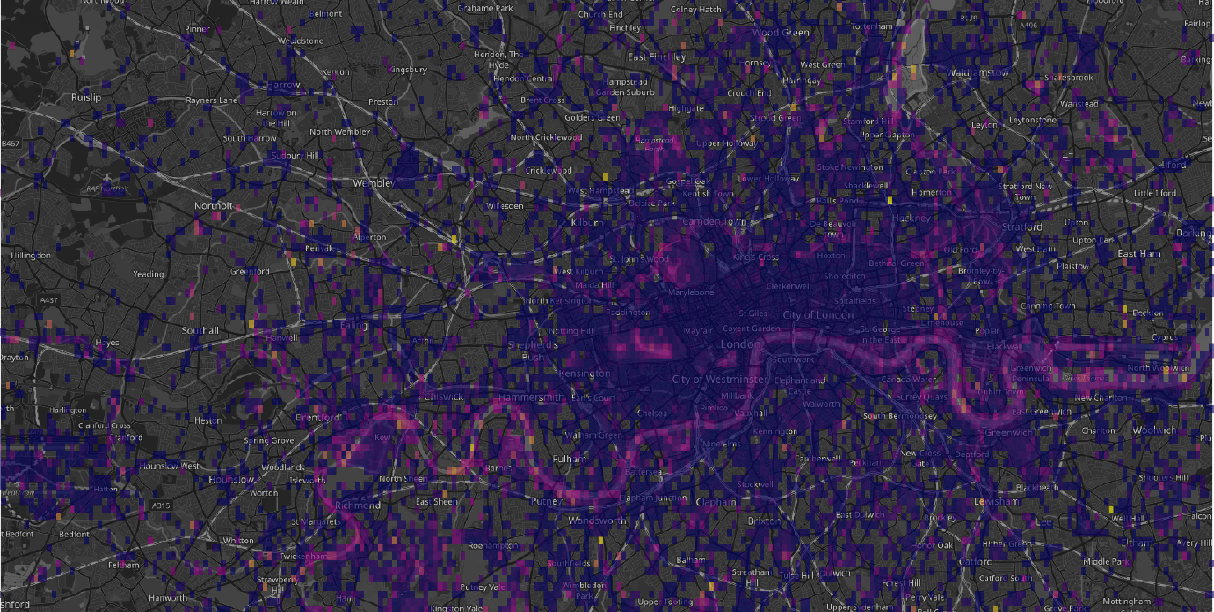}   &  \includegraphics[width=0.45\columnwidth]{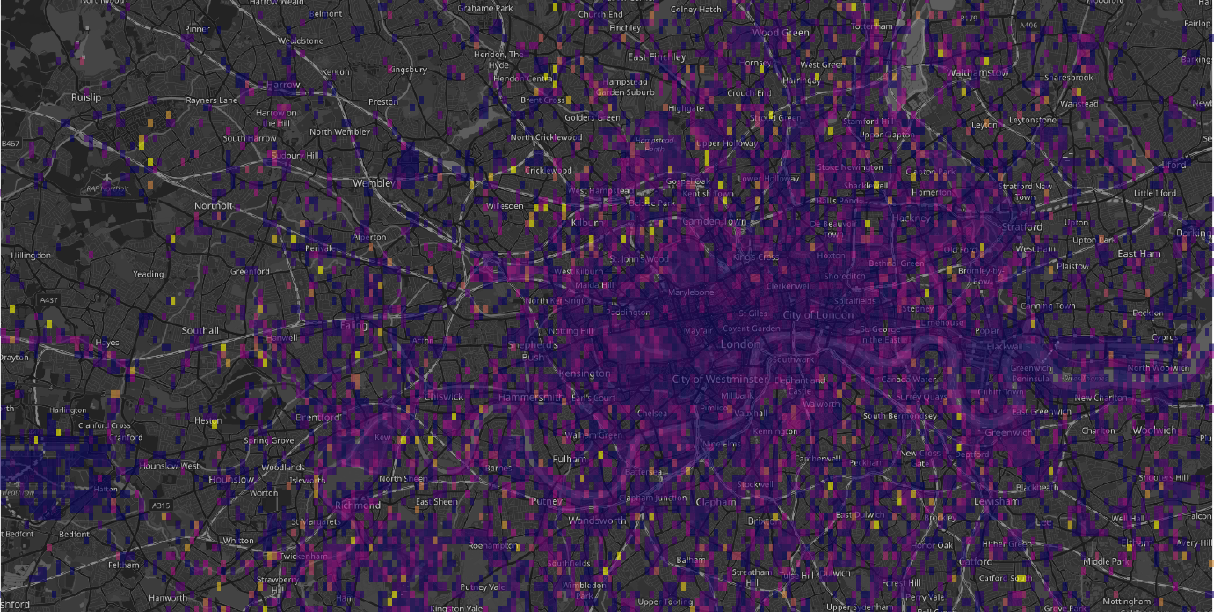}  \\
       \makecell{(c) vacationing/touring, climbing, \\diving, swimming, warm, natural (1.21)}  & \makecell{(d) shingles, brick, natural light,\\ man-made (-0.98)} \\
       \includegraphics[width=0.45\columnwidth]{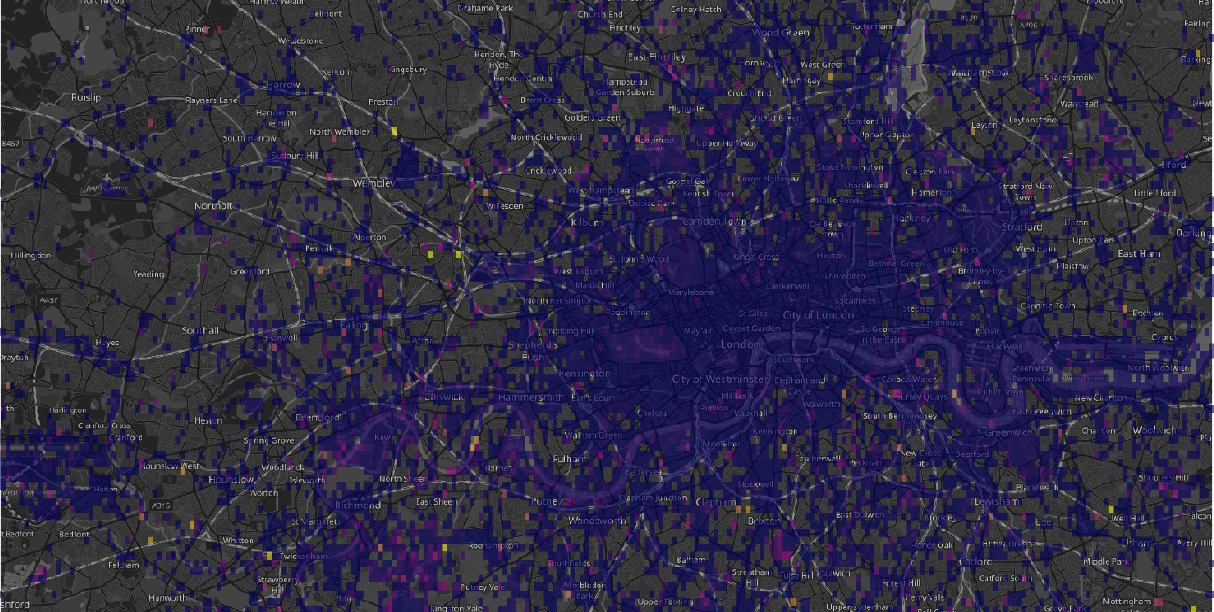}   &  \includegraphics[width=0.45\columnwidth]{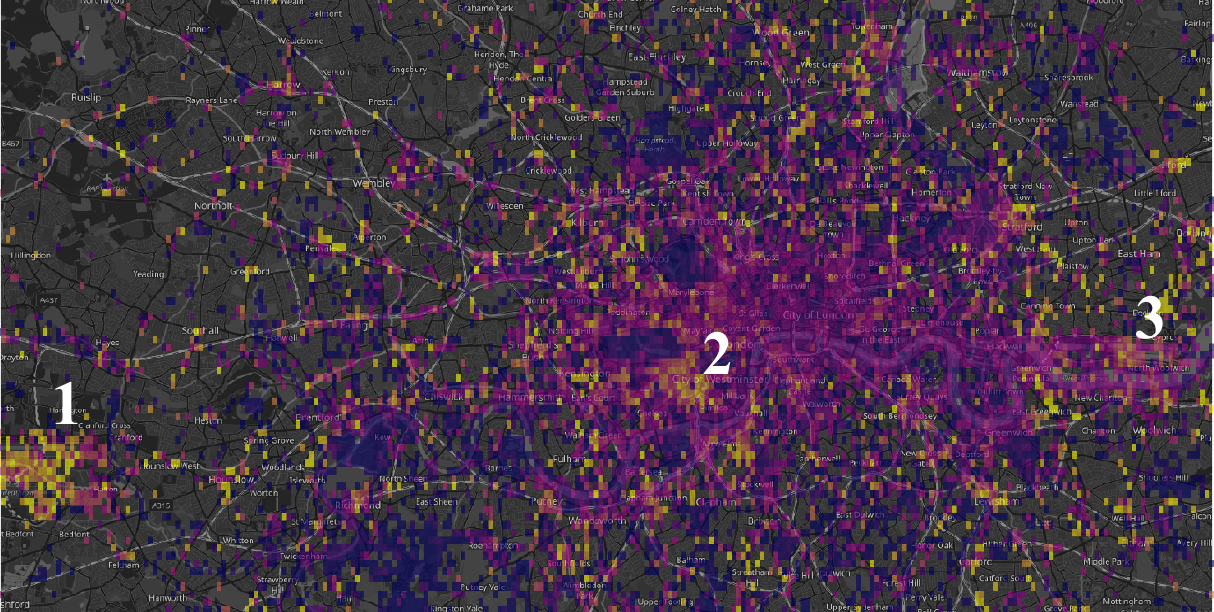}  \\
       \makecell{(e) climbing, rock/stone, ice, snow,\\ natural, rugged scene (0.86)}  & \makecell{(f) driving, transporting things or people,\\ asphalt, glossy, matte (-0.84)} \\
       \includegraphics[width=0.45\columnwidth]{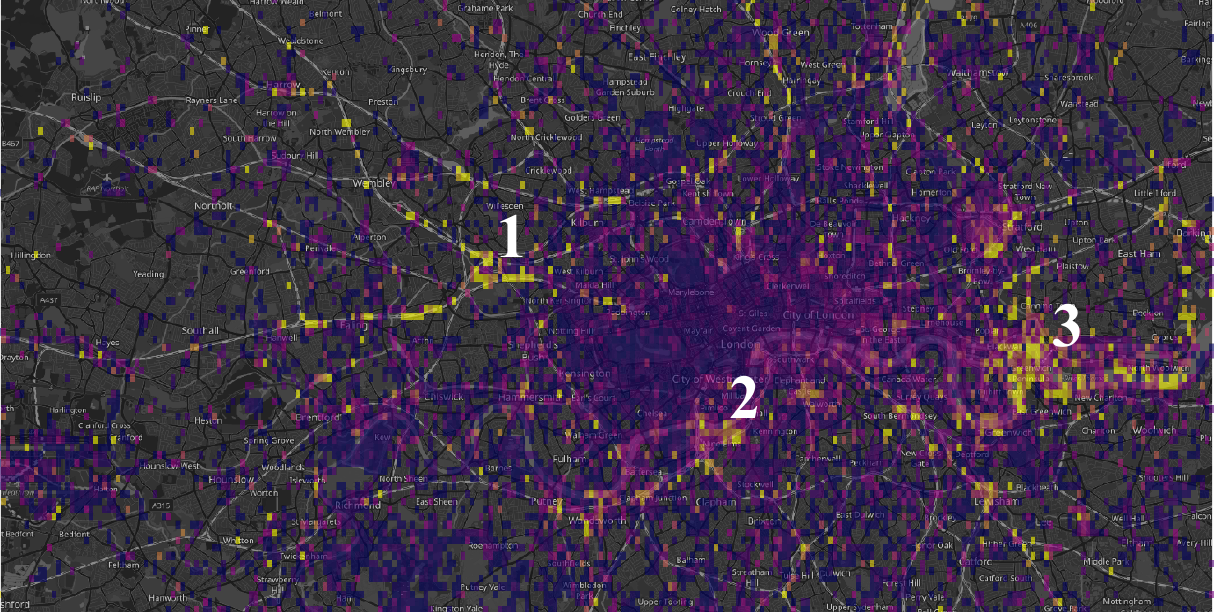}   &  \includegraphics[width=0.45\columnwidth]{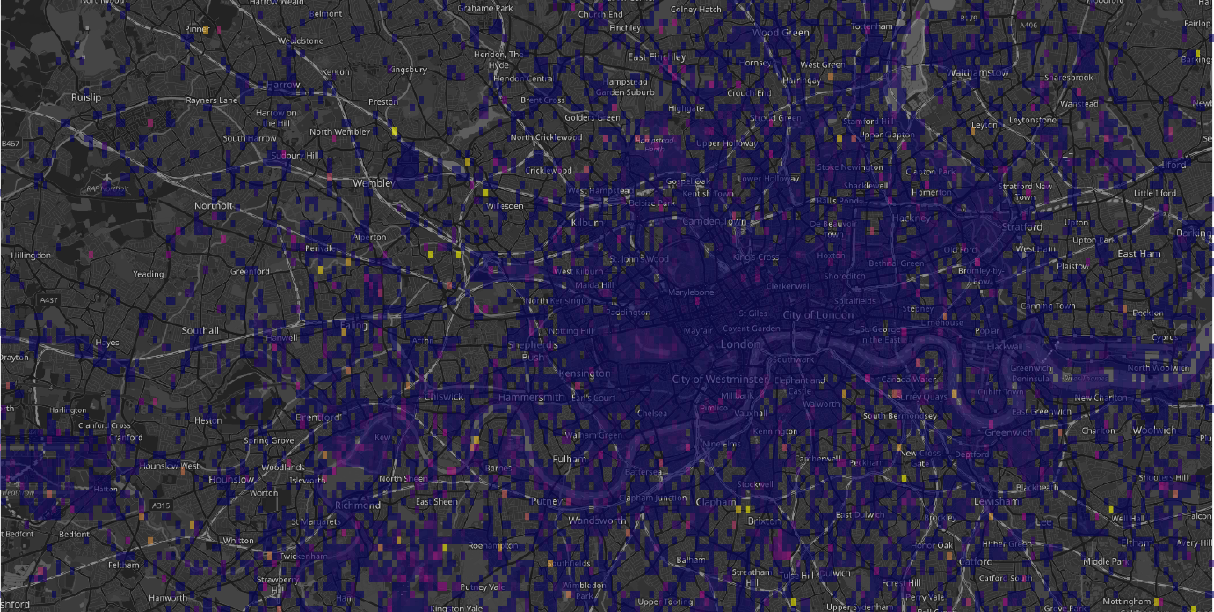}  \\
       \makecell{(g) constructing/ building, wire, metal,\\ rusty, man-made (-0.62)}  & \makecell{(h) climbing, ice, snow, cold (0.52)} \\
       \includegraphics[width=0.45\columnwidth]{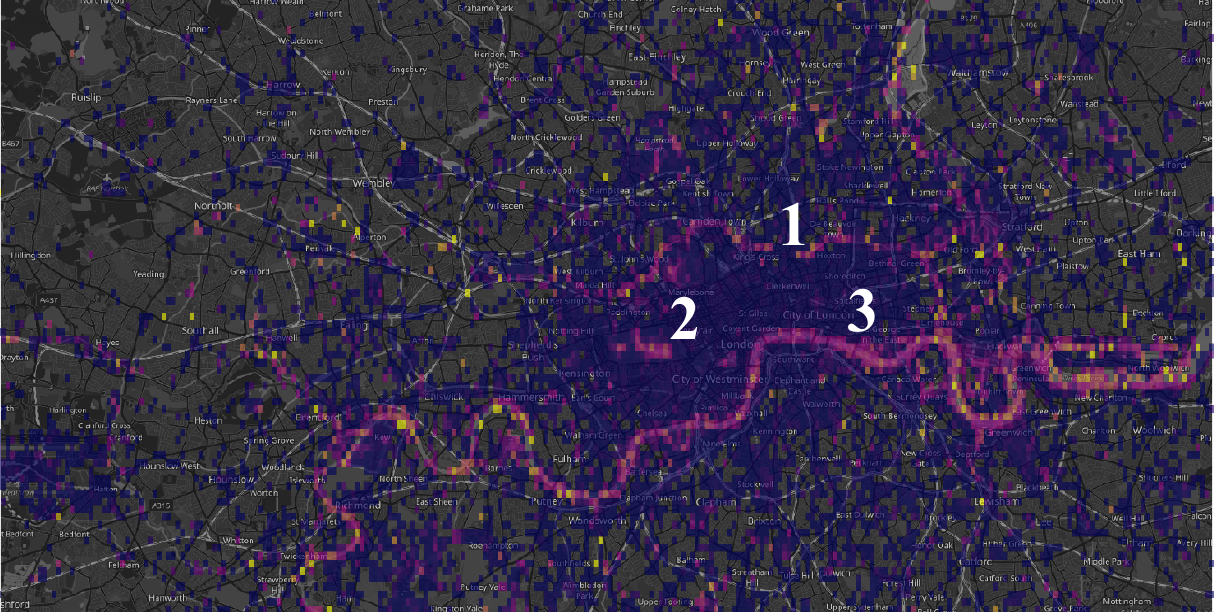}   &  \includegraphics[width=0.45\columnwidth]{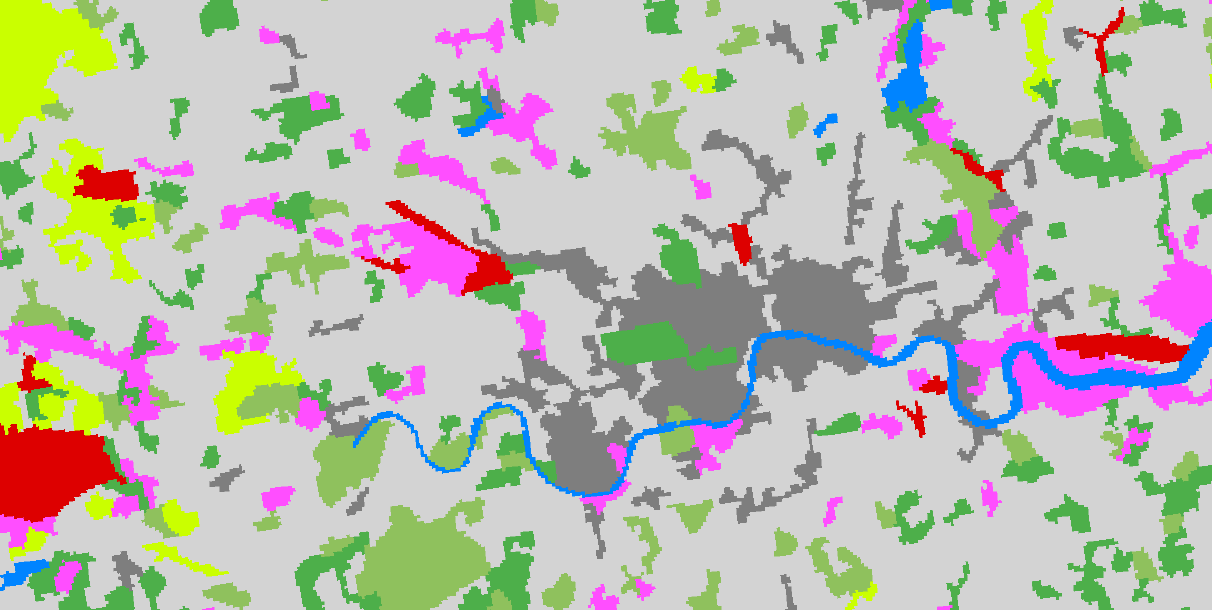}  \\
       \makecell{(i) diving, swimming, bathing,\\ waves/ surf, ocean (0.50)}  & (j)  \includegraphics[width=0.45\columnwidth]{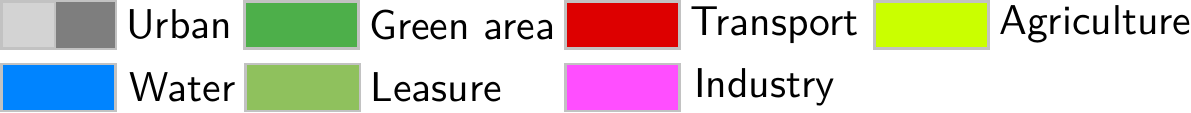}
    \end{tabular}
    
    \caption{CSBI group activation maps on the Flickr images in the London area (a - i), with color bar in (a), and CORINE land-cover map (j).}
    \label{fig:london_groups}
\end{figure}

Fig.~\ref{fig:scenicness} shows the scenicness maps computed with the Flickr images over London using the baseline and CSIB. Both maps show an overall agreement, with CSIB being visibly more conservative. Fig.~\ref{fig:london_groups} depicts the maps of CSIB group presence for the same area, showing how it allows, not just to map scenicness, but also to distinguish between different kinds thereof. For instance, unscenic areas dominated by group (g) are mostly related to industry and construction sites (g1 is to the area around Park Royal business park, g2 Nine Elms construction sites and g3 the Greenwich Peninsula), while those dominated by (f) tend to be associated to transport infrastructure (f1 and f3 correspond to the airports and f2 to Victoria station). At the same time, we are able to map elements, such as water surfaces (i1 is the Regent's canal, i2 the lake in Hyde park and i3 the Thames) even though their influence on scenicness is limited.

